\documentclass[sigconf]{acmart}

\usepackage{booktabs} 
\usepackage{multirow}
\usepackage{bm}
\usepackage{enumerate}
\usepackage{balance}
\usepackage{fancyhdr}

\setcopyright{rightsretained}

\acmArticle{4}
\acmPrice{15.00}

\begin{document}

\copyrightyear{2018}
\acmYear{2018}
\setcopyright{acmcopyright}
\acmConference[MM '18]{2018 ACM Multimedia Conference}{October 22--26, 2018}{Seoul, Republic of Korea}
\acmBooktitle{2018 ACM Multimedia Conference (MM '18), October 22--26, 2018, Seoul, Republic of Korea}
\acmPrice{15.00}
\acmDOI{10.1145/3240508.3240640}
\acmISBN{978-1-4503-5665-7/18/10}

\title{Decoupled Novel Object Captioner}

\author{Yu Wu$^{1}$, Linchao Zhu$^{1}$, Lu Jiang$^{2}$, Yi Yang$^{1,3}$}
\affiliation{%
  \institution{$^{1}$CAI, University of Technology Sydney, $^{2}$Google Inc. \\
  $^{3}$ State Key Laboratory of Computer Science, Institute of Software, Chinese Academy of Sciences}
	}
\email{yu.wu-3@student.uts.edu.au, yi.yang@uts.edu.au}

\def\ie{\emph{i.e.}}
\def\eg{\emph{e.g.}}
\def\etal{\emph{et~al.}}
\def\Pset{\mathcal{P}}
\def\USset{\mathcal{U_s}}
\def\WSset{\mathcal{W_s}}
\def\WUset{\mathcal{W_{unseen}}}
\def\NSet{\mathcal{N}}
\def\s{{\bf{s}}}
\def\h{{\bf{h}}}
\def\l{{\bf{l}}}
\def\L{{\bf{L}}}
\def\v{{\bf{v}}}
\def\f{{\bf{f}}}
\def\F{{\bf{F}}}
\def\w{{\bf{w}}}
\def\W{{\bf{W}}}
\def\Wset{{\mathcal{W}}}

\def\PL{\texttt{<PL>}}
\def\obj{\texttt{obj}}
\def\Mem{{\bm{\mathcal{M}}}_{\texttt{obj}}}

\def\I{{\bf{I}}}
\def\x{{\bf{x}}}

\begin{abstract}
Image captioning is a challenging task where the machine automatically describes an image by sentences or phrases. It often requires a large number of paired image-sentence annotations for training. However, a pre-trained captioning model can hardly be applied to a new domain in which some novel object categories exist, \ie, the objects and their description words are unseen during model training. To correctly caption the novel object, it requires professional human workers to annotate the images by sentences with the novel words. It is labor expensive and thus limits its usage in real-world applications.

In this paper, we introduce the zero-shot novel object captioning task where the machine generates descriptions without extra training sentences about the novel object. To tackle the challenging problem, we propose a Decoupled Novel Object Captioner (DNOC) framework that can fully decouple the language sequence model from the object descriptions.
DNOC has two components. 1) A \textbf{Sequence Model with the Placeholder} (SM-P) generates a sentence containing placeholders. The placeholder represents an unseen novel object. Thus, the sequence model can be decoupled from the novel object descriptions. 2) A \textbf{key-value object memory} built upon the freely available detection model, contains the visual information and the corresponding word for each object.
A query generated from the SM-P is used to retrieve the words from the object memory. The placeholder will further be filled with the correct word, resulting in a caption with novel object descriptions. The experimental results on the held-out MSCOCO dataset demonstrate the ability of DNOC in describing novel concepts in the zero-shot novel object captioning task.

\end{abstract}

%



\maketitle

\section{Introduction}
Image captioning is an important task in vision and language research \cite{karpathy2015deep,ranzato2015sequence,vinyals2015show,xu2015show}. It aims at automatically describing an image by natural language sentences or phrases. Recent encoder-decoder architectures have been successful in many image captioning tasks~\cite{bengio2015scheduled,donahue2015long,karpathy2015deep,mao2014deep,ranzato2015sequence,vinyals2015show,xu2015show}, in which the Convolutional Neural Network (CNN) is usually used as the image encoder, and the decoder is usually a Recurrent Neural Network (RNN) to sequentially predict the next word given the previous words. The captaining networks need a large number of image-sentence paired data to train a meaningful model.

\begin{figure}[t!]
	\centering
    \includegraphics[width=\linewidth]{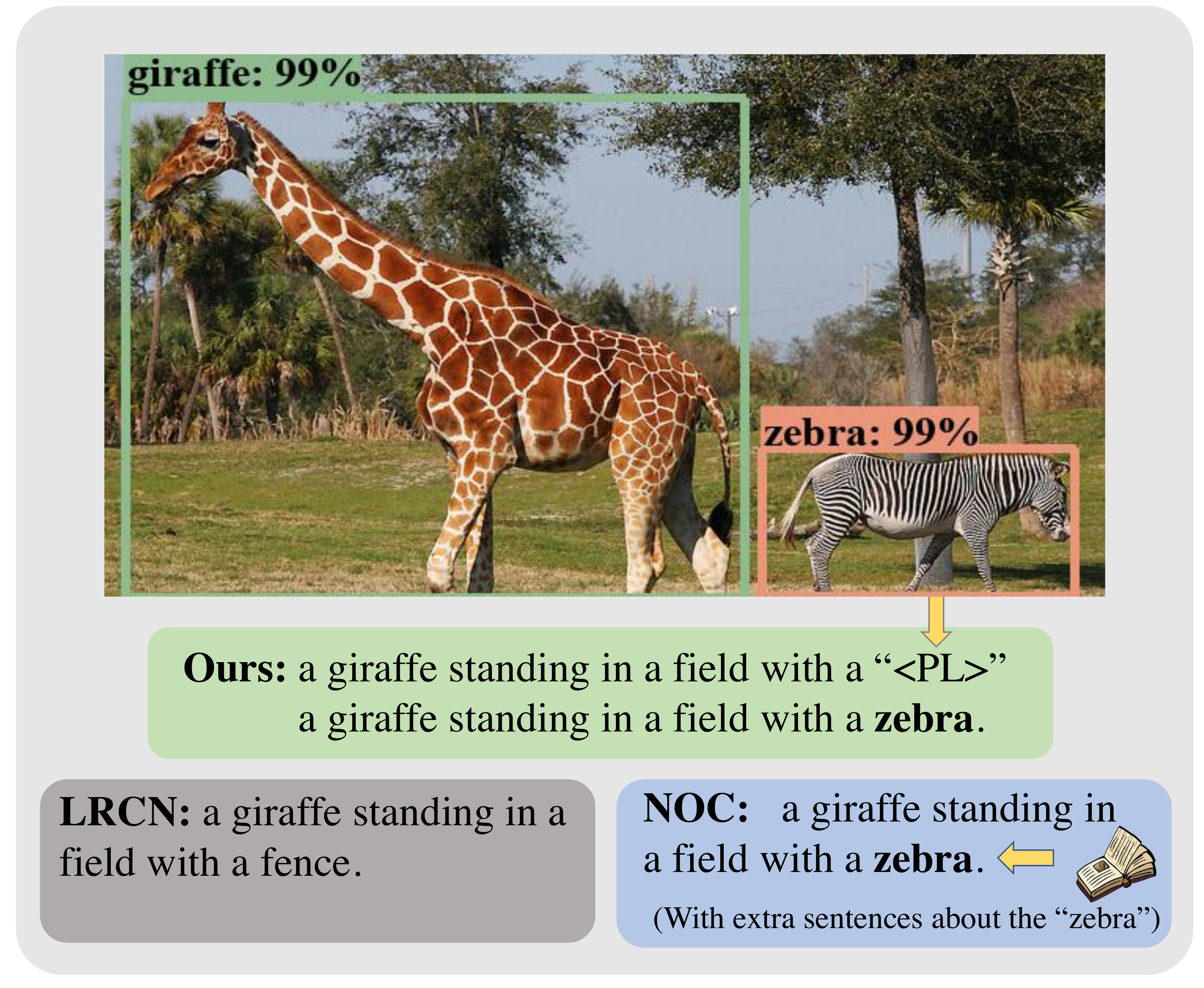}
    \vspace{-8mm}
    \caption{An example of the novel object captioning. The colored bounding boxes show the object detection results. The novel object ``zebra'' is not present in the training data.
    LRCN~\cite{donahue2015long} fails to describe the image with the novel object ``zebra''.
    NOC~\cite{venugopalan17cvpr} can generate the correct caption but requires extra text training data containing the word ``zebra'' to learn this concept.
    Our algorithm can generate correct captions, and \textit{more importantly}, we do not need any extra sentence data.
    Specifically, we first generate the caption template with a placeholder ``<PL>'' that represents the novel object. We then fill in the placeholder with the word ``zebra'' from the object detection model.
}
\label{fig:1task}
\vspace{-5mm}
\end{figure}

These captioning models fail in describing the \emph{novel objects} which are unseen words in the paired training data. For example, as shown in Figure \ref{fig:1task}, the LRCN \cite{donahue2015long} model cannot correctly generate captions for the novel object ``zebra''.
As a result, to apply the model in a new domain where novel objects can be visually detected, it requires professional annotators to caption new images in order to generate paired training sentences. This is labor expensive and thus limits the applications of captioning models.

A few works have been proposed recently to address the novel object captioning problem \cite{anne2016deep,venugopalan17cvpr,yao2017incorporating}. Essentially, these methods attempt to incorporate the class label produced by the pre-trained object recognition model.
The class label can be used as the novel object description in language generation. To be specific, Henzdricks~\etal~\cite{anne2016deep} trained a captioning model by leveraging a pre-trained image tagger model and a pre-trained language sequence model from external text corpora. 
Yao~\etal~\cite{yao2017incorporating} exploited a pre-trained sequence model to copy the object detection result into the output sentence.

However, to feed the novel object description into the generated captions, existing approaches either employ the pre-trained language sequence model \cite{anne2016deep,venugopalan17cvpr} or require extra unpaired training sentences of the novel object~\cite{yao2017incorporating}.
In both cases, the \textit{novel} objects have been used in training and, hence, is not really novel.
A more precise meaning of \textit{novel} in existing works is \textit{unseen} in the \textit{paired} training sentences. However, the word is \textit{seen} in the \textit{unpaired} training sentences.
For example, in Figure~\ref{fig:1task}, existing methods require extra training sentences containing the word ``zebra'' to produce the caption, even though the object zebra is confidently detected in the image.
The assumption that such training sentences of the novel object always exist may not hold in many real-world scenarios.
It is considerably difficult to collect sentences about brand new products in a timely manner, \eg, self-balancing scooters, robot vacuums, drones, and emerging topics trending on the social media like fidget spinners.
Someone may collect these training sentences about novel objects. However, it can still introduce language biases into the captioning model.
For example, if training sentences are all about bass (a sea fish), the captioning model will never learn to caption the instrument bass and may generate awkward sentences like ``A man is eating a bass with a guitar amplifier.''

In this paper, we tackle the image captioning for novel objects, where we do not need any training sentences containing the object. We utilize a pre-trained object detection model about the novel object.
We call it \textit{zero-shot} novel object captioning to distinguish it from the traditional problem setting~\cite{anne2016deep,venugopalan17cvpr,yao2017incorporating}.
In the traditional setting, in addition to the pre-trained object detection model, extra training sentences of the novel object are provided.
In the zero-shot novel object captioning, there are \textit{zero training sentences} about the novel object, \ie, there is no information about the semantic meaning, sense, and context of the object.
As a result, existing approaches of directly training the word embedding and sequence model become infeasible.

To address this problem, we propose a Decoupled Novel Object Captioner (DNOC) framework that is able to generate natural language descriptions without extra training sentences of the novel object.
DNOC follows the standard encoder-decoder architecture but with an improved decoder. Specifically, we first design a sequence model with the placeholder (SM-P) to generate captions with placeholders. The placeholder represents the unseen word for a novel object. Then we build a key-value object memory for each image, which contains the visual information and the corresponding words for objects.
Finally, a query is generated to retrieve a value from the key-value object memory and the placeholder is filled by the corresponding word.
In this way, the sequence model is fully decoupled from the novel object descriptions.
Our DNOC is thus capable of dealing with the unseen novel object. For example, in Fig.~\ref{fig:1task}, our method first generates the captioning sentence by generating a placeholder ``$<$PL$>$'' to represent any novel object. Then it learns to fill in the placeholder with ``zebra'' based on the visual object detection result. 

In summary, the main contributions of this work are listed as follows:
\begin{itemize}

\item We introduce the zero-shot novel object captioning task, which is an important research direction but has been largely ignored.
\item To tackle the challenging task, we design the sequence model with the placeholder (SM-P). SM-P generates a sentence with placeholders to fully decouple the sequence model from the novel object descriptions.

\item A key-value object memory is introduced to incorporate external visual knowledge. It interactively works with SM-P to tackle the zero-shot novel object captioning problem.
Extensive experimental results  on the held-out MSCOCO dataset show that our DNOC is effective for zero-shot novel object captioning. Without extra training data, our model significantly outperforms state-of-the-art methods (with additional training sentences) on the F1-score metric.

\end{itemize}

\section{Related Work}
The problem of generating natural language descriptions from visual data has been a popular research area in the computer vision field. 
This section first reviews recent works for image captioning and then extends the discussion to the zero-shot novel object captioning task.

\subsection{Image Captioning}
Image captioning aims at automatically describing the content of an image by a complete and natural sentence. This is a fundamental problem in artificial intelligence that connects computer vision and natural language processing \cite{show_and_tell_PAMI}. 
Some early works such as template-based approaches \cite{kulkarni2013babytalk,mitchell2012midge} and search-based approaches \cite{farhadi2010every,ordonez2011im2text} generate captioning by the sentence template and the sentence pool.
Recently, language-based models have achieved promising performance. Most of them are based on the encoder-decoder architecture to learn the probability distribution of both visual embedding and textual embeddings \cite{bengio2015scheduled,donahue2015long,Zhu2017,karpathy2015deep,kiros2014multimodal,mao2014deep,ranzato2015sequence,vinyals2015show,xu2015show,dong2018fpait}.
In this architecture, the encoder is a CNN model which processes and encodes the input image into an embedding representation, while the decoder is a RNN model that takes the CNN representation as the initial input and sequentially predicts the next word given the previous word. Among recent contributions, Kiros \etal~\cite{kiros2014multimodal} proposed a multi-modal log-bilinear neural language model to jointly learn word representations and image feature embeddings.
Vinyals \etal~\cite{vinyals2015show} proposed an end-to-end neural network consisting of a vision CNN followed by a language generating RNN.
Xu \etal~\cite{xu2015show} improved \cite{vinyals2015show} by incorporating the attention mechanism into captioning. 
The attention mechanism focuses on the salient image regions when generating corresponding words.
You \etal~\cite{you2016image} further utilized an independent high-level concepts/attribute detector to improve the attention mechanism for image captioning.
Tavakoliy~\etal~\cite{tavakoliy2017paying} proposed a saliency-boosted image captioning model in order to investigate benefits from low-level cues in language models.

\subsection{Zero-Shot Novel Object Captioning}
Zero-shot learning aims to recognize objects whose instances may not have been seen during training \cite{lampert2014attribute, jiang2015bridging, rohrbach2011evaluating, Zero_PAMI18, jiang2015fast}.
Zero-shot learning bridge the gap between the visual and the textual semantics by learning a dictionary of concept detectors on external data source. 

Novel object captioning is a challenging task where there is no paired visual-sentence data for the novel object in training. Only a few works have been proposed to address this captioning problem. Mao \etal~\cite{mao2015learning} tried to describe novel objects with only a few paired image-sentence data. 
Henzdricks \etal~\cite{anne2016deep} proposed the Deep Compositional Captioner (DCC), a pilot work to address the task of generating descriptions of novel objects which are not present in paired image-sentence datasets. 
DCC leverages a pre-trained image tagger model from large object recognition datasets and a pre-trained language sequence model from external text corpora.
The captioning model is trained on the paired image-sentence data with above pre-trained models.
Venugopalan~\etal~\cite{venugopalan17cvpr} discussed a Novel Object Captioner (NOC) to further improve the DCC to an end-to-end system by jointly training the visual classification model, language sequence model, and the captioning model. 
Anderson~\etal~\cite{anderson2016guided} leveraged an approximate search algorithm to forcibly guarantee the inclusion of selected words during the evaluation stage of a caption generation model.
Yao~\etal~\cite{yao2017incorporating} exploited a mechanism to copy the detection results to the output sentence with a pre-trained language sequence model.
Concurrently to us,  Lu~\etal~\cite{lu2018neural} also proposed to generate a sentence template with slot locations, which are then filled in by visual concepts from object detectors. However, when captioning the novel objects, they have to manually defined category mapping list to replace the novel object's word embedding with an existing one.

Note that all of the above methods have to use extra data of the novel object to train their word embedding. Different from existing methods, our method focuses on zero-shot novel object captioning task in which there are no additional sentences or pre-trained models to learn such embeddings for novel objects.
Our method thus needs to introduce a new approach to exploit the object word's meaning, sense, and embedding in the zero-shot training condition.

\section{The proposed Method}

In this section, we first introduce the preliminaries and further show the two key parts of the proposed Decoupled Novel Object Captioner,~\ie,~the sequence model with the placeholder (Section~\ref{sec:language_pl}) and the key-value object memory (Section~\ref{sec:key_value}).
The overview of the DNOC framework and training details are illustrated in Section~\ref{sec:dnoc_framework} and Section~\ref{sec:training}, respectively .

\subsection{Preliminaries}
We first introduce the notations for image captioning.
In image captioning, given an input image $I$, the goal is to generate an associated natural language sentence $\s$ of length $n_l$, denoted as $\s = (\w_1, \w_2, ..., \w_{n_l})$. 
Each $\w$ represents a word and the length $n_l$ is usually varied for different sentences.
Let $\Pset = \{ (\I_{1}, \s_{1}), ..., (\I_{n_p}, \s_{n_p})\}$ be the set with $n_p$ image-sentences pairs.
The vocabulary of $\Pset$ is $\mathcal{W}_{paired} = \{ \w_{1}, \w_{2}, ..., \w_{N_t} \}$ which contains $N_t$ words.
Each word $\w_i \in \{0, 1\}^{N_t}$ is a one-hot (1 of $N_t$) encoding vector. 
The one-hot vector is then embedded into a $D_w$-dimensional real-valued vector $\x_i=\phi_w(\w_i) \in {\mathbb{R}}^{D_w} $.  
The embedding function $\phi_w(\cdot)$ is usually a trainable linear transformation $\x_i={\bf{T}}_w \w_i$, where
${\bf{T}}_w \in {\mathbb{R}}^{D_w \times N_t}$ is the embedding matrix.
The typical architecture for captioning is the encoder-decoder model.
In the followings, we show the encoding and the decoding procedures during the testing phase.

\textbf{The encoder.}
We obtain the representation for an input image $\I$ by $\phi_e (\I)$, where $\phi_e(\cdot)$ is the embedding function for encoder.
The function $\phi_e$ is usually an ImageNet pre-trained CNN model with the classification layer removed. It extracts the top-layer outputs as the visual features.

\textbf{The decoder.}
The decoder is a word-by-word sequence model designed to generate the sentence given the encoder outputs. 
In specific, at the first time step $t=0$, a special token $\w_0$ (<GO>) is the input to the sequence model, which indicates the start of the sentence.
At time step $t$, the decoder generates a word $\w_{t}$ given the visual content $\phi_{e}(\I)$ and previous words $(\w_{0}, ..., \w_{t-1})$.
Therefore, we formulate the probability of generating the sentence $\s$ as
\begin{align}
p({\bf{s}}|\I) = \textstyle{\prod_{t=1}^{n_l}} p(\w_t|\w_0,\ldots,\w_{t-1}, \phi_e(\I)).
\label{eq:decoder}
\end{align}

The  Long Short-Term Memory (LSTM)~\cite{hochreiter1997long} is commonly used as the decoder in visual captioning and natural language processing tasks \cite{yao2017incorporating,anne2016deep,venugopalan2015sequence}.
The core of the LSTM model is the memory cell, which encodes the knowledge of the input that has been observed at every time step.
There are three gates that modulate the memory cell updating, \ie, the input gate, the forget gate, and the output gate.
These three gates are all computed by the current input $\mathbf{x}_t$ and the previous hidden state $\mathbf{h}_{t-1}$. The input gate controls how the current input should be added to the memory cell; the forget gate is used to control what the cell should forget from the previous memory; the output gate controls whether the current memory cell should be passed as output. 
Given inputs $\mathbf{x}_t$, $\mathbf{h}_{t-1}$, we get the predicted output word $\mathbf{o}_t$ by updating the LSTM unit at time step $t$ as follows:
\begin{align}
\mathbf{o}_t, \mathbf{h}_t = \texttt{LSTM} (\mathbf{x}_t, \mathbf{h}_{t-1}).
\label{eq:lstm}
\end{align}

Combining Eqn \eqref{eq:decoder} and Eqn \eqref{eq:lstm}, at $t$-th time step, the previous word $\w_{t-1}$ is taken as the decoder input $\mathbf{x}_t$. 
In the training stage, we feed the ground-truth word of the previous step as the model input. 
In the evaluation stage, we take the output $\mathbf{o}_{t-1}$ of the model at $(t-1)$-th time step as model input $\mathbf{x}_{t}$.

\textbf{Zero-Shot Novel Object Captioning.}
This paper studies the zero-shot novel object captioning task, where the model needs to caption novel objects \emph{without} additional training sentence data about the object.
The novel object words are shown neither in the paired image-sentence training data $\Pset$ nor unpaired sentence training data. 
We denote ${\mathcal {W}}_{unseen}$ as the vocabulary for the novel object words which are unseen in training. 
Given an input image $\I_{n}$ containing novel objects, the captioning model should generate a sentence with the corresponding unseen word $\tilde{\w} \in {\mathcal{W}}_{unseen}$ to describe the novel objects.

A notable challenge for this task is to deal with the out-of-vocabulary (OOV) words.
The learned word embedding function $\phi_w$ is unable to encode the unseen words, since these word cannot simply be found in $\Wset_{paired}$.
As a result, these unseen words cannot be fed into the decoder for caption generation.
Previous works \cite{anne2016deep,venugopalan17cvpr,yao2017incorporating} circumvented this problem by learning the word embeddings of unseen words using additional sentences that contain the words. We denote these extra training sentences as ${\mathcal{S}}_{unpaired}$. Since the words of ``novel'' objects have been used in training, the ``novel'' objects are not really novel. However, in our zero-shot novel object task, we do not assume the availability of additional training sentences ${\mathcal{S}}_{unpaired}$ of the novel object. Therefore, we propose a novel approach to deal with the OOV words in the sequence model.

\subsection{Sequence Model with the Placeholder}
\label{sec:language_pl}

\begin{figure}[t!]
	\centering
    \includegraphics[width=\linewidth]{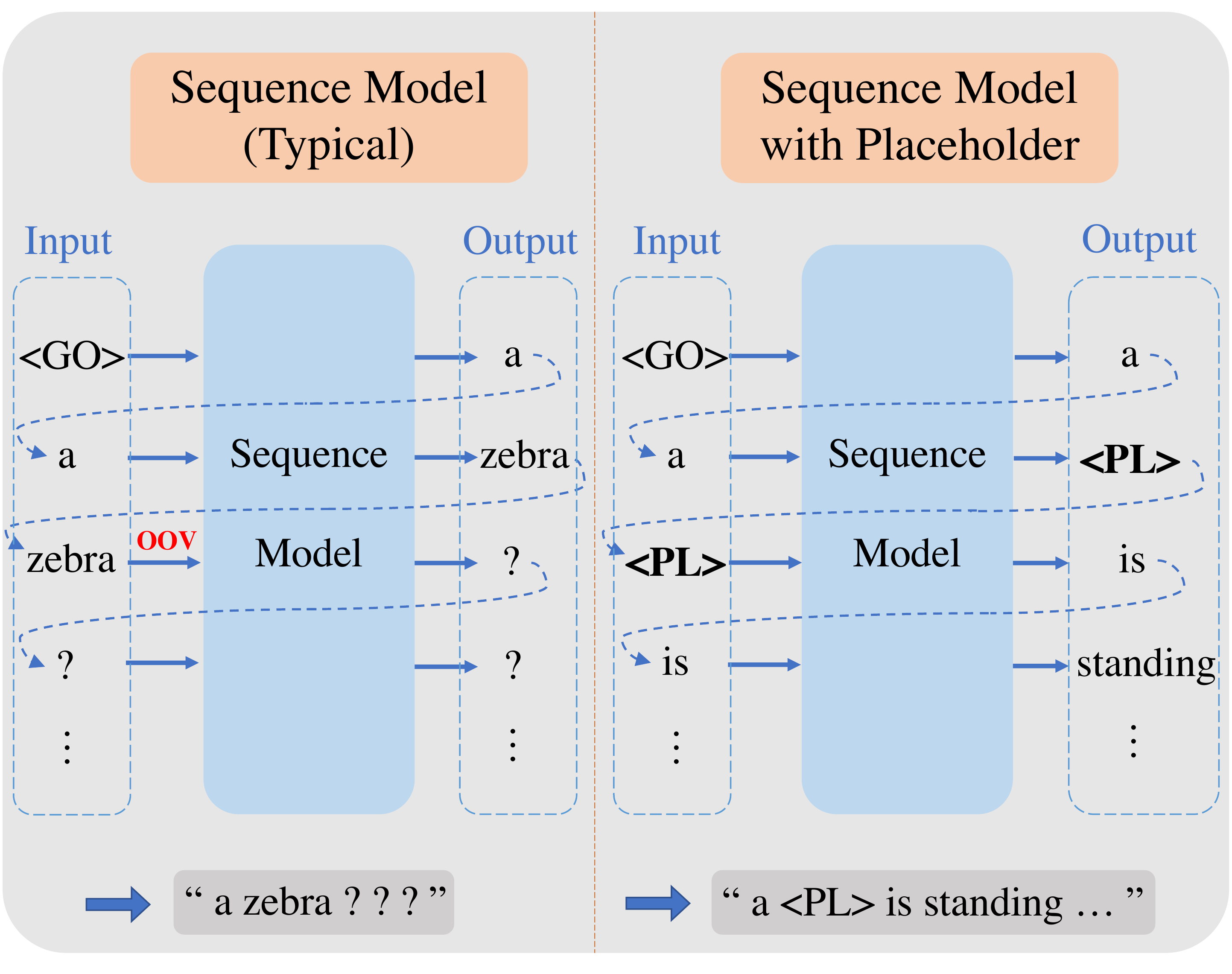}
    \vspace{-5mm}
    \caption{
    The comparison of the typical sequence model and the proposed SM-P.
    In this example, ``zebra'' is an unseen word during training. 
    The bottom are the sentences generated by the two models.
    The left is the classical sequence model, which cannot handle the input out-of-vocabulary (OOV) word ``zebra''.
    The right is our sequence model with the placeholder (SM-P).
    It generates the special token word ``<PL>'' (placeholder) to represent the novel object, and is able to continue to output the subsequent word given the input ``<PL>''.
    }
    \label{fig:2SM-P}
    \vspace{-5mm}
\end{figure}

We propose the Sequence Model with the placeholder (SM-P) to fully decouple the sequence model from novel object descriptions.
As discussed above, the classical sequence model cannot take an out-of-vocabulary word as input.
To solve this problem, we design a new token, denoted as ``<PL>''. ``<PL>'' is the \emph{placeholder} that represents any novel words $\tilde{\w} \in \Wset_{unseen}$.
It is used in the decoder similarly to other tokens, such as ``<GO>'', ``<PAD>'', ``<EOS>'', ``<UNKNOWN>'' in most natural language processing works. 
We add the token ``<PL>'' into the paired vocabulary $\Wset_{paired}$ to learn the embedding.
The training details for the placeholder are discussed in Section \ref{sec:training}.

A new embedding $\phi_w$ is learned for ``<PL>'', which encodes all unknown words with a compact representation. The new representation could be jointly learned with known words. 
We carefully designed the new token ``<PL>'' in both the input and the output of the decoder,
which enables us to handle the out-of-vocabulary words.
When the decoder outputs ``<PL>'', our model utilizes the external knowledge from the object detection model to replace it with novel description.
Our SM-P is flexible that can be readily incorporated in the sequence to sequence model.
Without loss of generality, we use the LSTM as the backbone of our SM-P.
When the SM-P decides to generate a word about a novel object at time step $t$, it will output a special word $\w_t$, the token ``<PL>'' , as the output.
At time step $t+1$, the SM-P takes the previous output word $\w_t$ ``<PL>'' as input instead of the novel word $\tilde{\w}$.
In this way, regardless of the existence of unseen words, the word embedding function $\phi_w$ is able to encode all the input tokens.
For example, in Figure \ref{fig:2SM-P}, the classical sequence model cannot handle the out-of-vocabulary word ``zebra'' as input. 
Instead, the SM-P model outputs the ``<PL>'' token when it needs to generate a word about the novel object ``zebra''.
This token is further fed to the decoder at the next time step. Thus, the subsequent words can be generated.
Finally, the SM-P generates the sentence with the placeholder ``A <PL> is standing ...''.
The ``<PL>'' token will be replaced by the novel word generated by the key-value object memory.

\begin{figure*}[t!]
    \begin{center}
    \includegraphics[width=\textwidth]{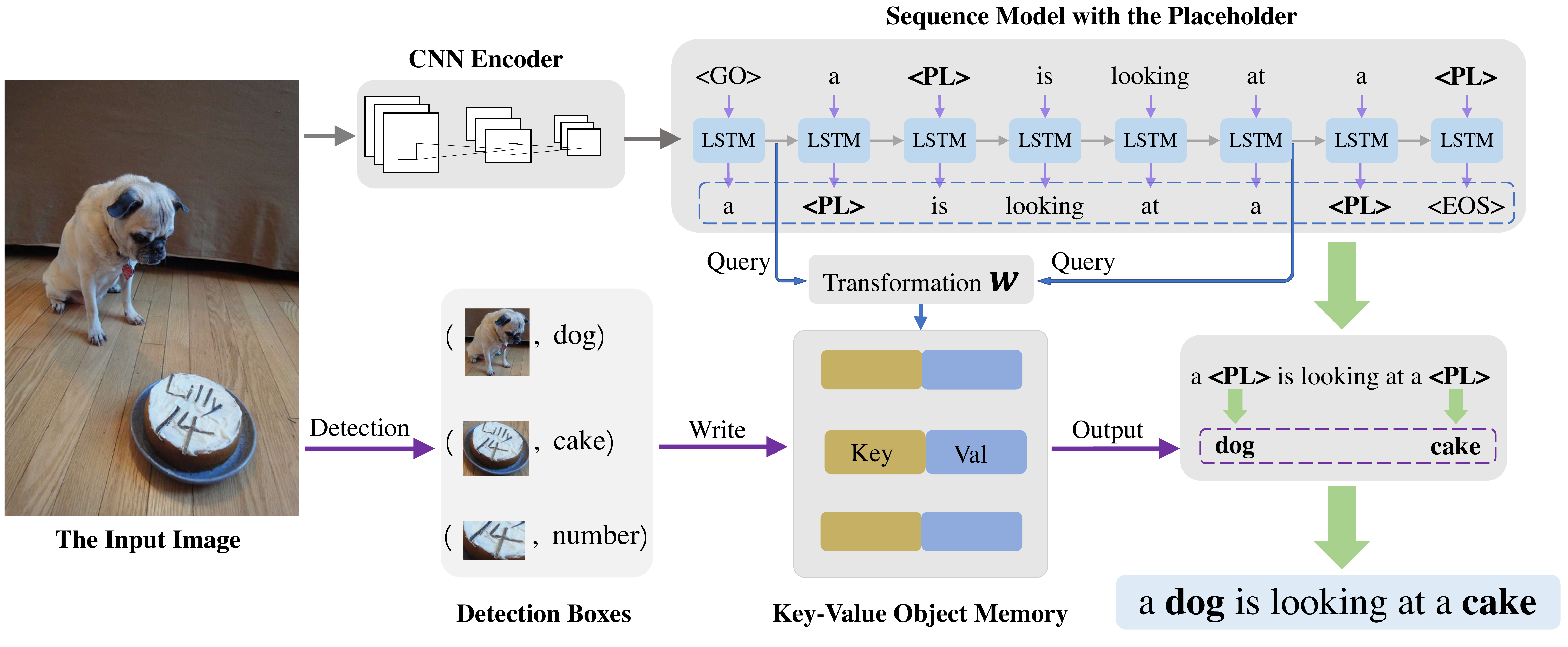}
    \end{center}
    \vspace{-4mm}
    \caption{The overview of the DNOC framework.
    We design a novel sequence model with the placeholder (SM-P) to handle the unseen objects by replacing them with the special token ``<PL>''.
    The SM-P first generates a sentence with placeholders, which refer to the unknown objects in an image.
    For example, in this figure, the ``dog'' and ``cake'' are unseen in the training set.
    The SM-P generates the sentence ``a <PL> is looking at a <PL>''.
    Meanwhile, we exploit a freely available object detection model to build a key-value object memory, which associate the semantic class labels (descriptions of the novel objects) with their appearance feature.
    When SM-P generates a placeholder, we take the linear transformation of the previous hidden state as a query to read the memory and output the correct object description, \eg, ``dog'' and ``cake''. Finally, we replace the placeholders by the query results and generate the sentence with novel words.
    }
    \label{fig:3framework}
    \vspace{-3mm}
\end{figure*}

\subsection{Key-Value Object Memory}
\label{sec:key_value}

To incorporate the novel words into the generated sentences with the placeholder, we exploit a pre-trained object detection model to build the key-value object memory.

A freely available object detection model is applied on the input images to find novel objects.
For the $i$-th detected object $\texttt{obj}_i$, the CNN feature representations $\f_i  \in \mathbb{R}^{1 \times N_f}$ and the predicted semantic class label $\l_i \in \mathbb{R}^{1 \times N_{D}}$ form a key-value pair, with the feature as key and the semantic label as the value. 
$N_f$ is the feature dimension of CNN representation, $N_{D}$ is the number of total detection classes.
The key-value pairs associate the semantic class labels (descriptions of the novel objects) with their appearance feature.
Following \cite{johnson2016densecap}, we extract the CNN feature $\f_i$ for $\texttt{obj}_i$ from the ROI pooling layer of the detection model.
Among all the detected results, the top $N_{det}$ key-value pairs are selected according to their confidence scores, which form the key-value object memory $\Mem$.
For each input image, the memory $\Mem$ is initialized to be empty. 

Let $\Wset_{det} $ be the vocabulary of the detection model, which consists of $N_{D}$ detection class labels.
Note that each word in $\Wset_{det}$ is the detection class label in the \emph{one-hot} format, since we cannot obtain trained word embedding function $\phi_w(\cdot)$ for the new word.
To generate the novel word and replace the placeholder in the sentence at time step $t$, we define the query $\bm{q}_{t}$ to be a linear transformation of previous hidden state $\h_{t-1}$ when the model meets the special token ``<PL>'' at time step $t$:
\begin{equation}
\bm{q}_{t} = w \h_{t-1},
\end{equation}
where $\h_{t-1} \in \mathbb{R}^{N_h}$ is the previous hidden state at $(t-1)$-th step from the sequence model, and $w \in \mathbb{R}^{N_f \times {N_h}}$ is the linear transformation to convert the hidden state from semantic feature space to CNN appearance feature space.
We have the following operations on the key-value memory $\Mem$:
\begin{align}
\Mem &\gets \texttt{WRITE} (\Mem, (\f_i, \l_i )) ,\\
\w_{\texttt{obj}} &\gets \texttt{READ} (\bm{q}, \Mem) . 
\end{align}

\begin{itemize}
\item \verb|WRITE| operation is to write the input key-value pair $(\f_i, \l_i)$ into the existing memory $\Mem$.
The input key-value pair is written to a new slot of the memory.
The key-value object memory is similar to the support set widely used in many few-shot works~\cite{vinyals2016matching,santoro2016one,finn2017model}. 
The difference is that in their work the key-value memory is utilized for long-term memorization, while our motivation is to build a structured mapping from the detection bounding-box-level feature to its semantic label.

\item \verb|READ| operation takes the query $\bm{q}$ as input, and conducts content-based addressing on the object memory $\Mem$. It aims to find related object information according to the similarity metric, $\bm{q} \bm{K}^T$.
The output of \verb|READ| operation is, 
\begin{equation}
\w_{\obj} = (\bm{q} \bm{K}^T) \bm{V},
\end{equation}
where $\bm{K}^T \in \mathbb{R}^{{N_f} \times {N_{det}}}, \bm{V} \in \mathbb{R}^{{N_{det}} \times {N_{D}}}$ are the vertical concatenations of all keys and values in the memory, respectively.
The output $\w_{\obj} \in \mathbb{R}^{N_{D}}$ is the combination of all semantic labels. 
In evaluation, the word with the max prediction is used as the query result.
\end{itemize}

\subsection{Framework Overview}
\label{sec:dnoc_framework}

With the above two components, we propose the DNOC framework to caption images with novel objects. 
The framework is based on the encoder-decoder architecture with the SM-P and key-value object memory.
For an input image with novel objects, we have the following steps to generate the captioning sentence:
\begin{enumerate}[(i)]
\item  We first exploit the SM-P to generate a captioning sentence with some placeholders.
Each placeholder represents an unseen word/phrase for a novel object;

\item  We then build a key-value object memory $\Mem$ for each input based on the detection feature-label pairs $\{\f_i, \l_i \}$ on the image;

\item  Finally, we replace the placeholders of the sentence by corresponding object descriptions. For the placeholder generated at time step $t$, we take the previous hidden state $\h_{t-1}$ from SM-P as a query to read the object memory $\Mem$, and replace the placeholder by the query results $\w_{obj}$.
\end{enumerate}
In the example shown in Figure \ref{fig:3framework}, the ``dog'' and ``cake'' are the novel objects which are not present in training. The SM-P first generates a sentence ``a <PL> is looking at a <PL>''. Meanwhile, we build the key-value object memory $\Mem$ based on the detection results, which contains both the visual information and the corresponding word (the detection class label).
The hidden state at the step before each placeholder is used as the query to read from the memory.
The memory will then return the correct object description, \ie, ``dog'' and ``cake''.
Finally, we replace the placeholders by the query results and thus generate the sentence with novel words ``a dog is looking at a cake''.

\subsection{Training}
\label{sec:training}
To learn how to exploit the ``out-of-vocabulary'' words, we modify the input and target for SM-P in training.
We define $\Wset_{pd}$ as the intersection set of the vocabulary $\Wset_{paired}$ and vocabulary $\Wset_{det}$,
\begin{equation}
\Wset_{pd} = \Wset_{paired} \cap \Wset_{det} .
\end{equation}

\noindent $\Wset_{pd}$ contains the words in the paired visual-sentence training data and the labels of the pre-trained detection model.
For the $i$-th paired visual-sentence input $(\I_{i}, \s_{i})$, we first encode the visual input by the encoder $\phi_{e}(\cdot)$.
We modify the input annotation sentence $\s_{i} = (\w_{1}, \w_{2}, ..., \w_{n_l})$ of the sequence model SM-P by replacing each word $\w_{i} \in \Wset_{pd}$ with the token ``<PL>''. 
The new input word $\hat{\w}_t$ at $t$-th time step is,
\begin{equation}
\hat{\w}_t=\left\{
\begin{array}{rcl}
\langle PL \rangle,            & {\w_t \in \Wset_{pd}}\\
\w_{t},          & { otherwise} .\\
\end{array} \right.
\end{equation}

\noindent We replace some known objects with the placeholder to help the SM-P learn to output the placeholder token at the correct place, and train the key-value object memory to output the correct word given the query. The actual input sentence for the SM-P is,
\begin{equation}
\hat{\s}_i = (\hat{\w}_0, \hat{\w}_1, ..., \hat{\w}_{n_l}) .
\end{equation}

\noindent We take the $\hat{\s}_i$ as the optimizing target for SM-P.
Let $F_{SM}(\cdot)$ denote the function of SM-P, and $\bm{\theta}_{SM-P}$ denote its parameters. 
The output of function $F_{SM}(\cdot)$ is the probability of next word prediction.
Each step of word generation is a word classification on existing vocabulary $\Wset_{paired}$.
SM-P is trained to predict the next word $\hat{\w}_t$ given $\phi_e(\I)$ and sequence of words $(\hat{\w}_0, \hat{\w}_1, ..., \hat{\w}_{t-1})$. 
The optimizing loss for $\mathcal{L}_{SM-P}$ is,
\begin{align}
\begin{split}
 &\mathcal{L}_{SM-P}(\hat{\w}_0, \hat{\w}_1,...,\hat{\w}_{t-1}, \phi_{e}(\I); \bm{\theta}_{SM-P}) = \\
& - \sum\limits_{t} \log(\texttt{softmax}_{t}({F_{SM}(\hat{\w}_0, \hat{\w}_1,...,\hat{\w}_{t-1}, \phi_{e}(\I); \bm{\theta}_{SM-P})}))  ,
\end{split}
\end{align}

\noindent where the $\texttt{softmax}_{t}$ denotes the softmax operation on the $t$-th step.

For the key-value object memory $\Mem$, we define the optimizing loss by comparing the query result $\w_{obj_{t}}$ from object memory and the word $\w_{t}$ from annotation, 
\begin{equation}
\label{eq:memory loss}
\mathcal{L}_{\Mem}= \sum\limits_{t} a_t CE(\w_{obj_{t}}, \w_t) ,
\end{equation}

\noindent where $CE(\cdot)$ is the cross-entropy loss function, and $a_{t}$ is the weight at time step~$t$ that is calculated by,
\begin{equation}
a_{t}=\left\{
\begin{array}{rcl}
1,            & {\w_t \in \Wset_{pd}}\\
0,          & \text{ otherwise} .\\
\end{array} \right.
\end{equation}

\noindent There are two trainable components in optimizing Eqn.~\eqref{eq:memory loss}. 
One is the query $\bm{q}$, the hidden state from the LSTM model. The other is the linear transformation on detection features in the computation of the memory key.
We simultaneously minimize the two loss functions. The final objective function for the DNOC framework is,
\begin{equation}
\mathcal{L}= \mathcal{L}_{SM-P} + \mathcal{L}_{\Mem} .
\end{equation}

\section{Experiments}
We first discuss the experimental setups and then compare DNOC with the state-of-the-art methods on the held-out MSCOCO dataset.
Ablation studies and qualitative results are provided to show the effectiveness of DNOC.

\subsection{The held-out MSCOCO dataset} \label{section:dataset}
The MSCOCO dataset \cite{lin2014microsoft} is a large scale image captioning dataset.
For each image, there are five human-annotated paired sentence descriptions.
Following~\cite{anne2016deep,venugopalan17cvpr,yao2017incorporating} , we employ the subset of the MSCOCO dataset, but excludes all image-sentence paired captioning annotations which describe at least one of eight MSCOCO objects.
The eight objects are chosen by clustering the vectors from the word2vec embeddings over all the 80 objects in MSCOCO segmentation challenge \cite{anne2016deep}. It results in the final eight novel objects for evaluation, which are "bottle", "bus", "couch", "microwave", "pizza", "racket", "suitcase", and "zebra". 
These novel objects are held-out in the training split and appear only in the evaluation split.
We use the same training, validation and test split as in \cite{anne2016deep}.

\subsection{Experimental Settings}
\noindent\textbf{The Object Detection Model}. We employ a freely available pre-trained object detection model to build the key-value object memory.
Specifically, we use Faster R-CNN \cite{ren2015faster} model with Inception-ResNet-V2 \cite{szegedy2017inception} to generate detection bounding boxes and scores. 
The object detection model is pre-trained on all the MSCOCO training images of 80 objects, including the eight novel objects.
We use the pre-trained models released by \cite{huang2017speed} which are publicly available.
For each image, we write the top $N_{det}=4$ detection results to the key-value object memory.

\begin{table*}[!tb]
\setlength{\tabcolsep}{3.5pt}
\centering
\caption{The comparison with the state-of-the-art methods on the eight novel objects in the held-out MSCOCO dataset. Per-object F1-score, averaged F1-score and METEOR score are reported.
All the results are reported using VGG-16~\cite{simonyan2014very} feature and without beam search.
Note that we adopt the zero-shot novel object captioning setting where no additional language data is used in training. With few training data, we achieve a higher average F1-score than all the previous methods with external sentence data.
All F1-score values are reported as percentage (\%).}
\label{table:FMCOCO}
\begin{tabular}{l|l|cccccccc|c|c}\hline
Settings&~~Methods&~F$_\text{bottle}$~&~F$_\text{bus}$~&~F$_\text{couch}$~&~F$_\text{microwave}$~&~F$_\text{pizza}$~&
~F$_\text{racket}$~&~F$_\text{suitcase}$~&~F$_\text{zebra}$~&~F$_\text{average}$~&~METEOR~
\\ \hline
\multirow{8}{*}{\shortstack{With\\External\\Semantic\\Data}}
                             &~~DCC \cite{anne2016deep}             & 4.63 & 29.79& 45.87 & 28.09 & 64.59& 52.24& 13.16& 79.88& 39.78 &21 \\
                             &~~NOC \cite{venugopalan17cvpr}         & & & &&&&&&&\\
                             &~~~--(One hot)                           &16.52 &68.63 &42.57 &32.16 &67.07 &61.22 &31.18 &88.39 &50.97 &20.7\\
                             &~~~~--(One hot+Glove)                     &14.93 &68.96 &43.82 &37.89 &66.53 &65.87 &28.13 &88.66 &51.85 &20.7\\
                             &~~LSTM-C\cite{yao2017incorporating}    & & & &&&&&&&\\
                             &~~~~--(One hot)                           &29.07 &64.38 &26.01 &26.04 &75.57 &66.54 &55.54 &\textbf{92.03}     &54.40 &22\\
                             &~~~~--(One hot+Glove)                     &29.68 &74.42  &38.77  &27.81  &68.17  &\textbf{70.27}    &44.76  &91.4 &55.66& \textbf{23}\\ 
                             &~~NBT+G~\cite{venugopalan17cvpr}         &7.1 &73.7 &34.4 &\textbf{61.9} &59.9 &20.2 &42.3 &88.5 &48.5 &22.8\\
\hline
\multirow{3}{*}{Zero-shot} &~~LRCN \cite{donahue2015long}         &0 &0 &0 &0 &0 &0 &0 &0 &0 &19.33\\
                      &~~DNOC w/o object memory                     &26.91 &57.14  &46.05  &41.88  &58.50 &18.41  &48.04   &75.17 &46.51 &20.41\\ 
                      &~~\textbf{DNOC (ours)}                 &\textbf{33.04} &\textbf{76.87} &\textbf{53.97} &46.57 &\textbf{75.82} &32.98 &\textbf{59.48} &84.58 &\textbf{57.92} &21.57\\ \hline
\end{tabular}
\end{table*}

\begin{table}[!tb]
\setlength{\tabcolsep}{2pt}
\centering
\caption{The comparison of our method and the baseline LRCN~\cite{donahue2015long} on the six known objects in the held-out MSCOCO dataset. Per-object F1-scors and averaged F1-scores are reported as percentage (\%).}
\label{table:Known}
\begin{tabular}{l|cccccc|c}\hline
Methods &F$_\text{bear}$ &F$_\text{cat}$& F$_\text{dog}$ &F$_\text{elephant}$& F$_\text{horse}$ &F$_\text{motorcycle}$ &F$_\text{average}$ \\
\hline
LRCN \cite{donahue2015long} &66.23  &75.73  &53.62  &65.49 &55.20 &71.45 &64.62 \\ 
\hline
DNOC &62.86 &87.28 &71.57 &77.46 &71.20 &77.59 &74.66\\
\hline
\end{tabular}
\vspace{-2mm}
\end{table}

\noindent\textbf{Evaluation Metrics}.
Metric for Evaluation of Translation with Explicit Ordering (METEOR)~\cite{banerjee2005meteor} is an effective machine translation metric which relies on the use of stemmers, WordNet \cite{miller1990introduction} synonyms and paraphrase tables to identify matches between candidate sentence and reference sentences. 
However, as pointed in \cite{anne2016deep,yao2017incorporating,venugopalan17cvpr}, the METEOR metric is not well designed for the novel object captioning task.
It is possible to achieve high METEOR scores even without mentioning the novel objects.
Therefore, to better evaluate the description quality, we also use the F1-score as an evaluation metric following \cite{anne2016deep,yao2017incorporating,venugopalan17cvpr}.
F1-score considers false positives, false negatives, and true positives, indicating whether a generated sentence includes a new object.

\noindent\textbf{Implementation Details}. Following~\cite{yao2017incorporating,anne2016deep,venugopalan17cvpr}, 
we use a 16-layer VGG \cite{simonyan2014very} pre-trained on the ImageNet ILSVRC12 dataset \cite{mitchell2012midge} as the visual encoder.
The CNN encoder is fixed during model training.
The decoder is an LSTM with cell size 1,024 and 15 sequence steps.
For each input image, we take the output of the fc7 layer from the pre-trained VGG-16 model with 4,096 dimensions as the image representation.
The representations are processed by a fully-connected layer and then fed to the decoder SM-P as the initial state.
For the word embedding, unlike \cite{yao2017incorporating,anne2016deep}, we do not exploit the per-trained word embeddings with additional knowledge data.
Instead, we learn the word embedding $\phi_{w}$ with 1,024 dimensions for all words including the placeholder token.
We implement our DNOC model with TensorFlow \cite{abadi2016tensorflow}.
Our DNOC is optimized by ADAM~\cite{kingma2014adam} with the learning rate of $1 \times 10^{-3}$. 
The weight decay is set to $5 \times 10^{-5}$.
We train the DNOC for 50 epochs and choose the model with the best validation performance for testing.

\subsection{Comparison to the state-of-the-art results}
We compare our DNOC with the following state-of-the-art methods on the held-out MSCOCO dataset.

(1). \textit{Long-term Recurrent Convolutional Networks (LRCN)}  \cite{donahue2015long}. LRCN is one of the basic RNN-based image captioning models. Since it has no mechanism to deal with novel objects, we train LRCN only on the paired visual-sentence data.

(2). \textit{Deep Compositional Captioner (DCC)} \cite{anne2016deep}. DCC leverages a pre-trained image tagger model from large object recognition datasets and a pre-trained language sequence model from external text corpora. The captioning model is trained on the paired image-sentence data with the two pre-trained models.

(3). \textit{Novel Object Captioner (NOC)} \cite{venugopalan17cvpr}. NOC improves the DCC to an end-to-end system by jointly training the visual classification model, language sequence model, and the captioning model. 

(4). \textit{LSTM-C} \cite{yao2017incorporating}. LSTM-C leverages a copying mechanism to copy the detection results to the output sentence with a pre-trained language sequence model.

(5). \textit{Neural Baby Talk (NBT)} \cite{lu2018neural}. NBT incorporates visual concepts from object detectors to the sentence template. They manually define category mapping list to replace the novel object's word embedding with an existing one to incorporate the novel words.

Table \ref{table:FMCOCO} summarizes the F1 scores and METEOR scores of the above methods and our DNOC on the held-out MSCOCO dataset.
All the state-of-the-art methods except LRCN use additional semantic data containing the words of the eight novel objects.
Nevertheless, without external sentence data, our method achieves competitive performance to the state of the art.
Our model on average yields a higher F1-score than the best state-of-the-art result (57.92\% versus 55.66\%).
The improvement is significant considering our model uses no additional training sentences.
Our METEOR score is slightly worse than the LSTM-C with GloVe \cite{yao2017incorporating}.
One possible reason is that much more training sentences containing the novel words are used to learn the LSTM-C model.

Table \ref{table:Known} shows the comparison of our DNOC and the baseline method LRCN on the known objects in the held-out MSCOCO dataset. 
Our method achieves much higher F1-scores on the known objects than LRCN, indicating that DNOC also benefits the captioning on the known objects. 
DNOC enhances the ability of the model to generate sentences with the objects shown in the image.
The results strongly support the validity of the proposed model in both the novel objects and the known objects.

\subsection{Ablation Studies}
The ablation studies are designed to evaluate the effectiveness of each component in DNOC.

\begin{table}[!tb]
\centering
\caption{Ablation studies in terms of Averaged F1-score and METEOR score on the held-out MSCOCO.
``LRCN'' is the baseline method.
``DNOC w/o detection model'' indicates the DNOC framework without any detected objects as input.
``DNOC w/o object memory'' indicates the DNOC framework with SM-P but without the key-value object memory.}
\begin{tabular}{|l|c|c|}\hline
~~Model    &    ~F1$_\text{average}$    ~&~METEOR~
\\ \hline
~~LRCN \cite{donahue2015long} &0 &19.33\\ 
~~DNOC w/o detection model          &0 &17.52\\
~~DNOC w/o object memory       &46.51 &20.41\\ 
~~DNOC          				&57.92 &21.57\\ \hline
\end{tabular} \label{tab3:ablation}
\vspace{-1mm}
\end{table}

\begin{figure*}[t!]
    \begin{center}
    \includegraphics[width=\textwidth]{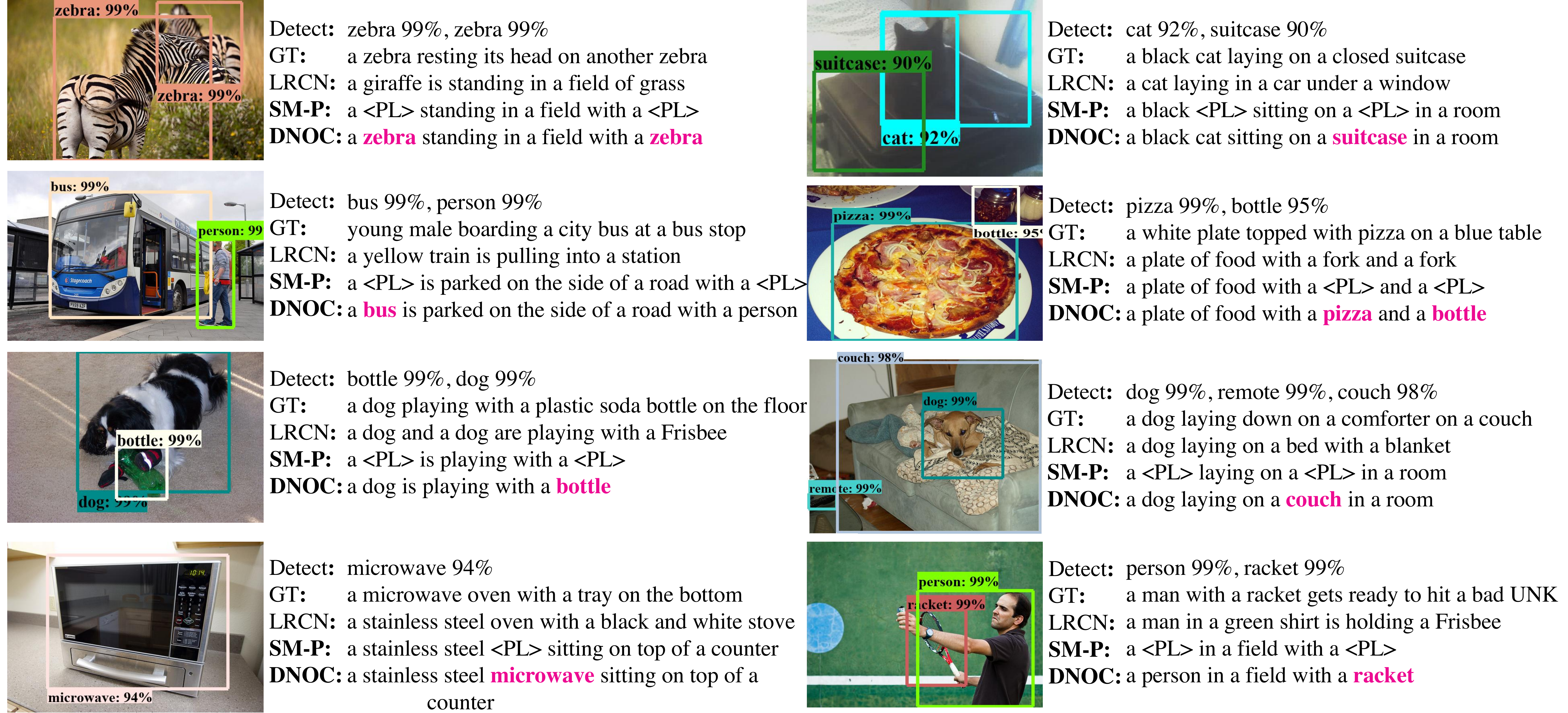}
    \end{center}
    \vspace{-3mm}
    \caption{
    Qualitative results for the held-out MSCOCO dataset.
    The words in pink are not present during training.
    ``Detected'' shows the object detection results.
    ``GT'' and ``LRCN'' are the human-annotated sentences and the sentences generated by LRCN, respectively.
    ``SM-P'' indicates the sentence generated by SM-P (the first step of DNOC).
    The SM-P first generates a sentence template with placeholder, and DNOC further feeds the detection results into the placeholder.
    }
    \label{fig:4visual}
    \vspace{-2mm}
\end{figure*}

\begin{figure}[t]
    \begin{center}
    \includegraphics[width=\linewidth]{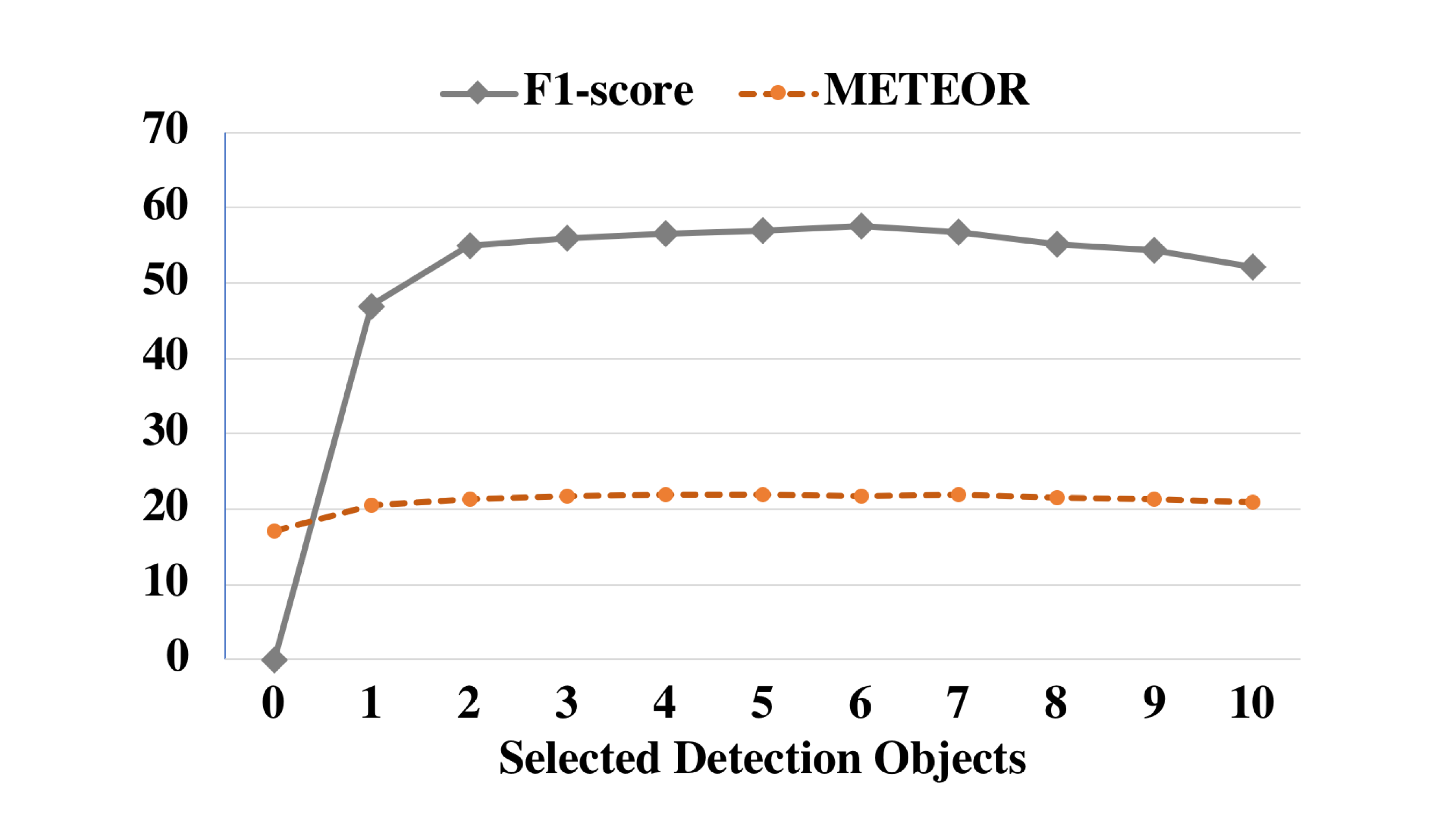}
    \end{center}
    \vspace{-5mm}
    \caption{The performance curve with different number of selected detected objects $N_{det}$.
    }
    \label{fig:ablation}
    \vspace{-4mm}
\end{figure}

\textbf{The effectiveness of SM-P and key-value object memory.}
We conduct the ablation studies on the held-out MSCOCO dataset. The results are shown in Table \ref{tab3:ablation}.
The ``DNOC w/o detection model'' indicates the DNOC framework with SM-P but without any detection objects as input.
All the objects in the visual inputs will not be detected. Thus, the placeholder token remains in the final generated sentence.
We observe a performance drop compared to LRCN.
``DNOC w/o object memory'' indicates the DNOC framework with SM-P but without the key-value object memory.
In this experiment, we take the top $N_{det}$ confident detected labels and randomly feed them into the placeholder.
Only with the SM-P component, ``DNOC w/o object memory'' outperforms LRCN by 46.51\% in F1-score and 1.08\% in METEOR score, which shows the effectiveness of the SM-P component.
Our full DNOC framework outperforms ``DNOC w/o object memory'' by 11.41\% in F1-score and 1.16\% in METEOR score.
It validates that the key-value object memory can enhance the semantic understanding of the visual content.
From the experimental results, we can conclude that our full DNOC framework with SM-P and the key-value object memory greatly improves the performance in both the F1-score and the METEOR score.
It shows that the two components are effective in exploiting the external detection knowledge.

\textbf{Analysis of the number of selected detection objects} $\bm{N_{det}}$.
$N_{det}$ is the number of selected top detection results.
We show the performance curves with different $N_{det}$ values in Figure \ref{fig:ablation}. When $N_{det}$ varies in a range from two to ten, we can see that the curves of F1-score and METEOR are relatively smooth.
When we only adopt one detected object to build the object memory $\Mem$, the F1-score significantly drops.
If we do not write any detection results into the memory, the F1-score is zero.
This curve also demonstrates the effectiveness of the key-value object memory.

\subsection{Qualitative Result}
In Figure \ref{fig:4visual}, we show some qualitative results on the held-out MSCOCO dataset.
We take the ``zebra'' image in the first row as an example.
The classic captioning model LRCN \cite{donahue2015long} could only describe the image with a wrong word ``giraffe'', where ``zebra'' does not existed in the vocabulary.
SM-P first generates the sentence with the placeholder, where each placeholder represents a novel object.
The DNOC then replaces the placeholders with the detection results by querying the key-value object memory. It generates the sentence with the novel word ``zebra'' in the correct place. 
As can be seen, zero-shot novel object captioning is a very challenging task since the evaluation examples contain unseen objects and no additional sentence data is available.

\section{Conclusions}
In this paper, we tackle the novel object captioning under a challenging condition where none sentence of the novel object is available. We propose a novel Decoupled Novel Object Captioner (DNOC) framework to generate natural language descriptions of the novel object. 
Our experiments validate its effectiveness on the held-out MSCOCO dataset. The comprehensive experimental results demonstrate that DNOC outperforms the state-of-the-art methods for captioning novel objects.

{\bf Acknowledgment.}
This work is partially supported by an Australian Research Council Discovery Project. We acknowledge the Data to Decisions CRC (D2D CRC) and the Cooperative Research Centres Programme for funding this research.

\newpage

\bibliographystyle{ACM-Reference-Format}
\balance
\bibliography{sample-bibliography}

\end{document}